\pdfoutput=1

\documentclass[11pt]{article}

\usepackage{acl}

\usepackage{times}
\usepackage{latexsym}

\usepackage[T1]{fontenc}
\usepackage[utf8]{inputenc}

\usepackage{microtype}

\usepackage{color}
\usepackage{adjustbox}
\usepackage{booktabs}
\usepackage{makecell}
\usepackage{amsfonts}
\usepackage{amsthm}
\usepackage{amsmath}
\usepackage{amssymb}
\usepackage{multirow}
\usepackage{xfrac}
\usepackage{scalerel}
\usepackage{float}
\usepackage{inconsolata} 
\usepackage[belowskip=-10pt,aboveskip=10pt]{caption}
\usepackage{subcaption}
\usepackage{scalerel,xparse}
\usepackage{graphicx}
\NewDocumentCommand\emojieyes{}{
    \includegraphics[scale=0.035]{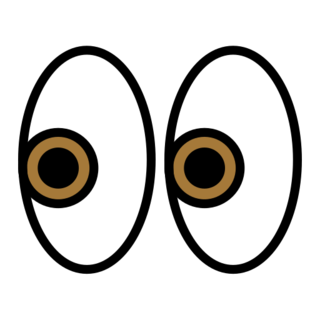}
}
\NewDocumentCommand\emojibook{}{
    \includegraphics[scale=0.04]{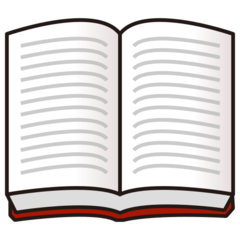}
}
\NewDocumentCommand\emojiglass{}{
    \includegraphics[scale=0.04]{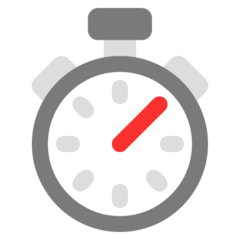}
}
\setcitestyle{notesep={ }}

\usepackage{cleveref}
\crefname{section}{\S}{\S\S}
\Crefname{section}{\S}{\S\S}
\crefname{table}{Tab.}{}
\crefname{figure}{Fig.}{Figs.}
\crefname{algorithm}{Algorithm}{}
\crefname{equation}{Eq.}{Eqs.}
\crefname{line}{Line}{}
\crefname{appendix}{App.}{}
\crefformat{section}{\S#2#1#3}
\crefname{thm}{Theorem}{}
\crefname{cor}{Corollary}{}
\crefname{prop}{Proposition}{}
\crefname{def}{Definition}{}

\definecolor{MacroColor}{RGB}{0,120,148}
\newcommand{\mymacro}[1]{{#1}}
\newcommand{\citeposs}[1]{\citeauthor{#1}'s (\citeyear{#1})}

\newcommand{\ww}{\mymacro{\boldsymbol w}}
\newcommand{\model}{\mymacro{q}}
\newcommand{\vocab}{\mymacro{\mathcal{V}}}

\newcommand{\infcontent}{\mymacro{\textsc{inf}}}
\newcommand{\infcontentk}{\mymacro{\textsc{inf}^{(k)}}}
\newcommand{\defeq}[0]{\mathrel{\stackrel{\textnormal{\tiny def}}{=}}}
\newcommand{\defn}[1]{\textbf{#1}}

\newcommand{\deltaLL}{\mymacro{\Delta\mathrm{LL}}}

\usepackage{todonotes}
\makeatletter
\newcommand*\iftodonotes{\if@todonotes@disabled\expandafter\@secondoftwo\else\expandafter\@firstoftwo\fi}  %
\makeatother

\title{Analyzing Wrap-Up Effects through an Information-Theoretic Lens}

\newcommand{\ucambridge}{\emojibook}
\newcommand{\ethz}{\emojieyes}
\newcommand{\MIT}{\emojiglass}
\usepackage{inconsolata}
\usepackage{color}

\author{Clara Meister$^{\ethz}$~\;~Tiago Pimentel$^{\ucambridge}$~\;~Thomas Hikaru Clark$^{\MIT}$\\
\textbf{Ryan Cotterell$^{\ethz}$~\;~Roger Levy$^{\MIT}$}\\
  $^{\ethz}$ETH Z\"{u}rich~\;~ $^{\ucambridge}$University of Cambridge~\;~ $^{\MIT}$Massachusetts Institute of Technology \\
  \texttt{\href{mailto:clara.meister@inf.ethz.ch}{clara.meister@inf.ethz.ch}}~\;~\texttt{\href{mailto:tp472@cam.ac.uk}{tp472@cam.ac.uk}}~\;~ \texttt{\href{mailto:thclark@mit.edu}{thclark@mit.edu}} \\
   \texttt{\href{mailto:ryan.cotterell@inf.ethz.ch}{ryan.cotterell@inf.ethz.ch}}~\;~ \texttt{\href{mailto:rplevy@mit.edu}{rplevy@mit.edu}}}

\begin{document}
\maketitle
\begin{abstract}
Numerous analyses of reading time (RT) data have been implemented---all in the effort to better understand the cognitive processes driving reading comprehension.  
However, data measured on words at the end of a sentence---or even at the end of a clause---is often omitted due to the confounding factors introduced by so-called ``wrap-up effects,'' which manifests as a skewed distribution of RTs for these words. 
Consequently, the community's understanding of the cognitive processes that might be involved in these wrap-up effects is limited.
In this work, we attempt to learn more about these processes by examining the relationship between wrap-up effects and information-theoretic quantities, such as word and context surprisals.  
We find that the distribution of information in prior contexts is often predictive of sentence- and clause-final RTs (while not of sentence-medial RTs). 
This lends support to several prior hypotheses about the processes involved in wrap-up effects.\looseness=-1 

\hspace{.5em}\includegraphics[width=1.25em,height=1.25em]{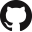}{\hspace{.75em}\parbox{\dimexpr\linewidth-2\fboxsep-2\fboxrule}{\footnotesize \url{https://github.com/rycolab/wrap-up-effects}}}
\end{abstract}

\section{Introduction}
Reading puts the unfolding of linguistic input in the hands---or, really, the eyes---of the reader. 
Consequently, it presents a unique opportunity to gain a better understanding of how humans comprehend written language. 
The rate at which humans choose to read text (and process its information) should be determined by their goal of understanding it. Ergo, examining where a reader spends their time should help us to understand the nature of language comprehension processes themselves.
Indeed, studies analyzing reading times have been employed to explore a number of psycholinguistic theories \citep[e.g.,][]{smith2013-log-reading-time,futrell2020lossy, van2021single}.\looseness=-1

One behavior revealed by such studies is the tendency for humans to spend more time\footnote{Longer reading times in self-paced reading studies and longer fixation times in eye-tracking studies.} 
on the last word of a sentence or clause. While the existence of such \defn{wrap-up effects} is well-known \cite{Just1982ParadigmsAP,hill2000,rayner2000,CAMBLIN2007103}, the cognitive processes giving rise to them are still not fully understood.
This is likely (at least in part) due to the dearth of analyses targeting naturalistic sentence-final reading behavior. 
First, most studies of online processing omit data from these words to explicitly control for the confounding factors wrap-up effects introduce \citep[e.g.,][]{smith2013-log-reading-time,goodkind-bicknell:2018predictive}.
Second, the few studies on wrap-up effects rely on small datasets, none of which analyze naturalistic text \cite{Just1980ATO,rayner2000,kuperberg2011}.   %
This work addresses this gap, using several large corpora of reading time data. Specifically, we study whether information-theoretic concepts (such as surprisal) provide insights into the cognitive processes that occur at a sentence's boundary.
Notedly, information-theoretic approaches have been proven effective for analyzing sentence-medial reading time behavior.\looseness=-1

We follow the long line of work that has connected information-theoretic measures and psychometric data \citep[, \emph{inter alia}]{frank2015erp,goodkind-bicknell:2018predictive,wilcox2020,meister+al.emnlp21a}, employing similar methods to build models of sentence- and clause-final RTs.
Using surprisal estimates from state-of-the-art language models, we search for a link between wrap-up effects and the information content within a sentence. 
We find that the distribution of surprisals of prior context is often predictive of sentence- and clause-final reading times (RTs), while not adding significant predictive power to models of 
sentence-medial RTs. This result suggests that the nature of cognitive processes involved during the reading of these boundary words may indeed be different than those at other positions. Such findings lend support to several prior hypotheses regarding which processes may underlie wrap-up effects (e.g., the resolution of prior ambiguities) while providing evidence against other speculations (e.g., that the time spent at sentence boundaries can be quantified with a constant factor, independent of the processing difficulty of the text itself).\looseness=-1%

\section{The Process of Reading}

Decades of research on reading behavior have improved our understanding of the cognitive processes involved in reading comprehension \cite[][, \emph{inter alia}]{Just1980ATO,RAYNER20094}. 
Here, we will briefly describe overarching themes that are relevant for understanding wrap-up effects.\looseness=-1

\subsection{Incrementality and its Implications}\label{sec:inc}

It is widely accepted that language processing is incremental, i.e., readers process text one word at a time \citep[][, \emph{inter alia}]{hale-2001-probabilistic,Hale2006UncertaintyAT,RAYNER20094,boston2011}.
Consequently, much can be uncovered about reading comprehension via studies that analyze cognitive processing at the word level. 
Many pyscholinguistic studies make use of this notion, taking per-word RTs in self-paced reading (SPR) or eye-tracking studies to be a direct reflection of the processing load of that word \citep[e.g.,][]{smith2013-log-reading-time,van2021single}.
This RT--processing effort relationship then allows us to identify relationships between a word's processing load and its attributes (e.g., surprisal or length)---which in turn hints at the underlying cognitive processes involved in comprehension. 
One prominently studied attribute is word predictability; a notion naturally quantified by \defn{surprisal} (also known as \citeposs{shannon1948mathematical} information content). 
Formally, the surprisal of a word $w$ is defined as $s(w) \defeq - \log p(w  \mid \ww_{< t})$, i.e., a unit's negative log-probability given the prior sentential context $\ww_{<t}$. 
Notedly, this operationalization provides a way of quantifying how our prior expectations can affect our ability to process a linguistic signal.

There are several hypotheses about the mathematical nature of the relationship between per-word surprisal and processing load.\footnote{Surprisal theory \cite{hale-2001-probabilistic}, for instance, posits a linear relation.\looseness=-1}
While there has been much empirical proof that surprisal estimates serve as a good predictor of word-level RTs \cite{smith2013-log-reading-time,goodkind-bicknell:2018predictive,wilcox2020}, the data observed from sentence-final words appears not to follow the same relationship. %
Specifically, in comparison to sentence-medial words, sentence- or clause-final words are associated with increased RTs in self-paced studies \cite{Just1982ParadigmsAP,hill2000} and both increased fixation and regression times in eye-tracking studies \cite{rayner2000,CAMBLIN2007103}. %
Such behavior has also been observed in controlled settings---for example, \citet{rayner1989} found that readers fixated longer on a word when it ended a clause than when the same word did not end a clause. 

Such widespread experimental evidence suggests sentence-final and sentence-medial reading behaviors differ from each other, and that other cognitive processes (besides standard word-level processing) effort may be at play. 
Yet unfortunately, these wrap-up effects have received relatively little attention in the psycholinguistic community: Most reading time studies simply exclude sentence-final (or even clause-final) words from their analyses, claiming that the (poorly understood) effects are confounding factors in understanding the reading process \citep[e.g.,][]{Frank2013ReadingTD,frank2015erp,wilcox2020}. Rather, we believe this data can potentially provide new insights in their own right.\looseness=-1%

\subsection{Wrap-up Effects}\label{sec:wrap}
It remains unclear what exactly occurs in the mind of the reader at the end of a sentence or clause. Which cognitive processes are encompassed by the term \defn{wrap-up effects}? Several theories have been posited.
First, \citet{Just1980ATO} hypothesize that wrap-up effects include actions such as ``the constructions of inter-clause relations.''
Second, \citet{rayner2000} suggest they might involve attempts to resolve previously postponed comprehension problems, which could have been deferred in the hope that upcoming words would resolve the problem. 
Third, \citet{HIROTANI2006425} posit the hesitation when crossing clause boundaries is out of efficiency \cite{JARVELLA1971409}; readers do not want to have to return to the clause later, so they take the extra time to make sure there are no inconsistencies in the prior text. 

While some prior hypotheses have been largely dismissed \citep[see ][ for a more detailed summary]{STOWE2018232} due to, e.g., the wide-spread support of theories of incremental processing, most others lack formal testing in 
naturalistic reading studies. %
We attempt to address this gap. Concretely, we posit the relationship between text's information-theoretic attributes and its observed wrap-up times can indicate the presence (or lack) of several cognitive processes that are potentially a part of sentence wrap-up.  
For example, high-surprisal words in the preceding context may correlate with the presence of ambiguities in the text; they may also correlate with complex linguistic relationships of the current text with prior sentences---which are two driving forces in the theories given above. 
Consequently, in this work, we ask whether the reading behavior observed at the end of a sentence or clause can be described (at least partially) by the distribution of information content in the preceding context,\footnote{Importantly, the research questions we ask are not concerned with describing the \emph{full} set of cognitive processes that occur at the end of a clause or sentence---or even whether there is a \emph{causal} relationship between information content and sentence- and clause-final RTs.} as this may give insights for several prior hypotheses about wrap-up effects.\looseness=-1 
\newcommand{\eos}{\mymacro{\textsc{eos}}}
\section{Language Models as Predictors of Psychometric Data}

Formally, a language model $\model$ is a probability distribution over natural language sentences, i.e.,
over $\vocab^*$ for an alphabet $\vocab$ of linguistic units (typically words).
In the case when $\model$ is locally normalized, which is the predominant case for today's neural language models, $\model$ is defined as the product of conditional probability distributions, i.e., for $\ww \in \vocab^*$, we have $\model(\ww) = \model(\eos \mid \ww) \prod_{t=1}^{|\ww|}\model(w_t \mid \ww_{<t})$, where each $\model(\cdot\! \mid \!\ww_{<t})$ is a distribution over $\vocab \cup \{\eos\}$.
The symbol $\eos$ is a special end-of-string token not in $\vocab$.
Consequently, we can use $\model$ to estimate the probability of individual words in context.
The model parameters are typically estimated by minimizing the negative log-likelihood of a corpus of natural language strings $\mathcal{C}$, i.e., minimizing $\mathcal{L}(\model) = -\sum_{\ww \in \mathcal{C}}\log \model(\ww)$ with respect to $\model$.\looseness=-1

One widely embraced technique in information-theoretic psycholinguistics is the use of these language models to estimate the probabilities required for computing surprisal
\cite{hale-2001-probabilistic,DEMBERG2008193,mitchell-etal-2010-syntactic,fernandez-monsalve-etal-2012-lexical}. %
It has even been observed that a language model's perplexity\footnote{Perplexity is a monotonic function of the average surprisal of linguistic units in-context under a model.\looseness=-1} correlates negatively with the psychometric predictive power provided by its surprisal estimates \cite{frank-bod,goodkind-bicknell:2018predictive,wilcox2020}. 
If these language models keep improving at their current fast pace \cite{radford2019language,brown2020language}, exciting new results in computational psycholinguistics may follow, connecting reading behavior to the statistics of natural language. 

\paragraph{Predicting Reading Times.} 
In the computational psycholinguistics literature, the RT--surprisal relationship is typically studied using predictive models: RTs are predicted using surprisal estimates (along with other attributes such as number of characters) for the current word. 
The predictive power of these models, together with the structure of the model itself (which defines a specific relationship between RTs and surprisal), is then used as evidence of the studied effect. 
While this paradigm is successful in modeling sentence-medial RTs \cite{smith2013-log-reading-time,goodkind-bicknell:2018predictive,wilcox2020}, its effectiveness for modeling sentence- and clause-final times is largely unknown due to the omission of this data from the majority of RT analyses.\looseness=-1

A priori, we might expect per-word surprisal to be a similarly powerful predictor of sentence and clause-final RTs.\footnote{Several authors \citep[e.g.,][]{STOWE2018232} have argued the cognitive processes involved in the comprehension of clause-final words are the same as those for sentence-medial words.} 
\begin{figure}
         \centering
         \includegraphics[height=0.5\linewidth]{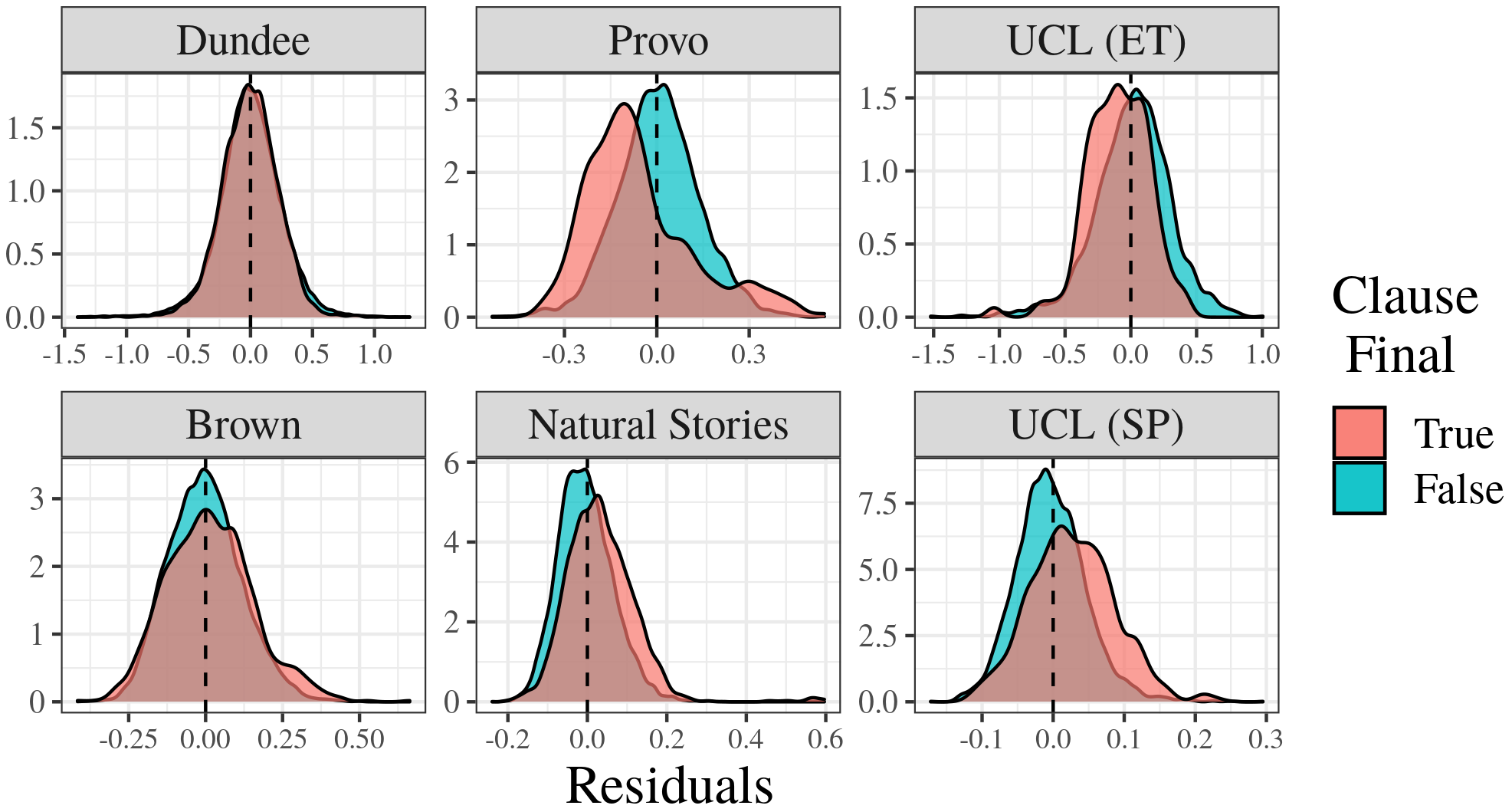}
        \caption{Distributions of %
        residuals when predicting either clause-final or non-clause-final times using our baseline linear models. Models are fit to (the log-transform of) non-clause-final average RTs.  Outlier times (according to log-normal distribution) are excluded. %
        The top-level datasets contain eye-tracking data while the bottom contain SPR data. Full distributions of RTs are shown in \cref{app:results}, where we also show models fit to regression times, rather than full reading times.\looseness=-1}
        \label{fig:densities}
\end{figure}
Yet in \cref{fig:densities}, we see that when our baseline linear model (described more precisely in \cref{sec:exps}) is fit to sentence-medial RTs, the residuals for predictions of clause-final RTs appear to be neither normally distributed nor centered around 0. Further, these trends appear to be different for eye-tracking and SPR data, where the latter are skewed towards \emph{lower} values for all datasets.\footnote{The opposite is true for regression times in eye-tracking data; see \cref{app:results}.}  
These results provide further confirmation that 
clause-final data does not adhere to the same relationship with RT as sentence-medial data, a phenomenon that may perhaps be accounted for by additional factors at play in the comprehension of clause-final words. 
Thus, we ask whether taking into account information from the entire prior context can give us a better model of these clause-final RTs.

\begin{figure*}
{\small (a)}
     \begin{subfigure}{0.89\textwidth}
         \includegraphics[width=\textwidth,trim={0 0 3cm 0},clip]{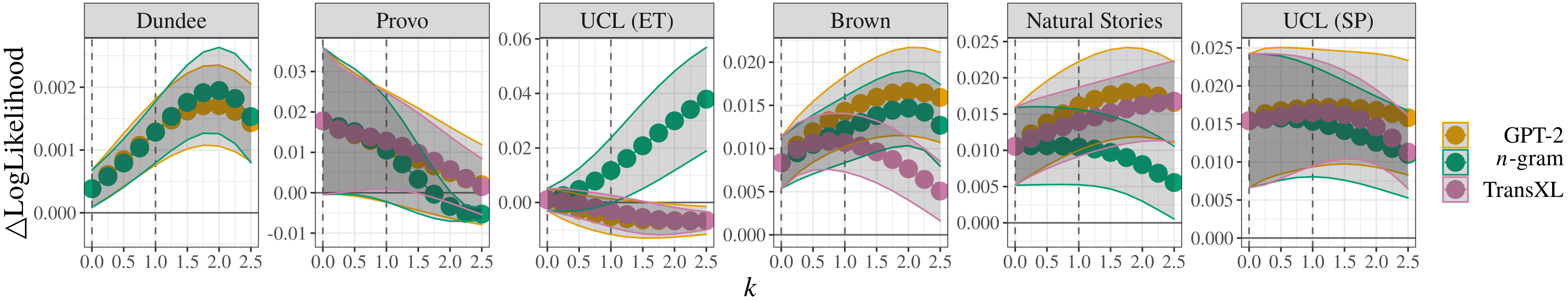}
     \end{subfigure}
      \\ {\small (b)}
     \begin{subfigure}{0.97\textwidth}
         \includegraphics[width=\textwidth]{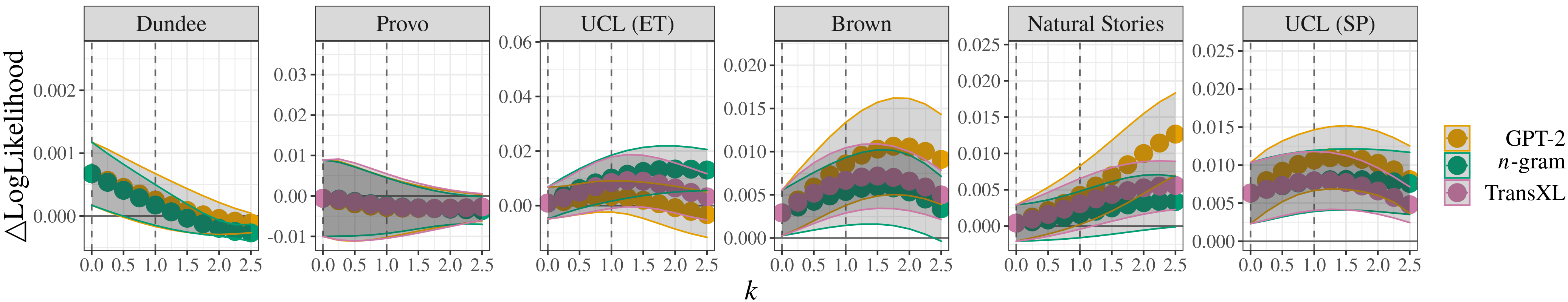}
     \end{subfigure}
    \caption{Mean $\deltaLL$ as a function of the exponent $k$ in $\infcontentk$ for models of sentence and clause-final (top row) and sentence-medial (bottom row) RTs using surprisal estimates from different language models. The shaded region connects standard error estimates. Vertical intercepts at $k=0,1$ are for reference. We see that our information-theoretic predictors contribute much less modeling power to the prediction of sentence-medial RTs in comparison to sentence- and clause-final RTs.  }
    \label{fig:LL_full}
\end{figure*}

To this end, we operationalize the information content $\infcontent$ in text $\ww$ (of length $T$) as:\footnote{We note \citet{meister+al.emnlp21a} used similar operationalizations to test for evidence in support of the uniform information density hypothesis.}\looseness=-1 
\begin{equation}\label{eq:power}
    \textstyle\infcontentk(\ww) \defeq \sum_{t=1}^{T} s(w_{t})^k \quad\quad {\color{gray} (k \geq 0)}
\end{equation} 
where $\ww$ may be an entire sentence or only its first $T$ words.\footnote{\color{blue} We omit $\eos$ from \cref{eq:power} for brevity.}
Notably, the case of $k=0$ returns $T$; under $k = 1$, we get the total information content of $\ww$. For $k>1$, moments of high surprisal will disproportionately drive up the value of $\infcontentk(\ww)$. Such words may indicate, e.g., moments of ambiguity or uneven distributions of information in text. Thus, how well $ \infcontentk(\ww)$ (as a function of $k$) predicts model sentence- and clause-final RTs may indicate which attributes of prior text (if any) can be linked to the additional cognitive processes involved in wrap-up effects.\looseness=-1

\section{Experiments}\label{sec:exps}

\paragraph{Data.} We use reading time data from 5 corpora over 2 modalities: the Natural Stories \cite{futrell-etal-2018-natural2}, Brown \cite{smith2013-log-reading-time}, and UCL (SP) \cite{Frank2013ReadingTD} Corpora, which contain SPR data, as well as the Provo \cite{provo}, Dundee \cite{dundee} and UCL (ET) \cite{Frank2013ReadingTD} Corpora, which contain eye movements during reading. All corpora are in English. For eye-tracking data, we take reading time to be the sum of all fixation times on that word. We provide an analysis of regression (a.k.a. go-past) time in \cref{app:results}. %
We provide further details regarding pre-processing in \cref{app:data}.

\paragraph{Estimating Surprisal.} We obtain surprisal estimates from three language models: GPT-2 \cite{radford2019language}, TransformerXL \cite{dai-etal-2019-transformer} and a 5-gram model, estimated using Modified Kneser--Essen--Ney Smoothing \cite{NEY19941}. We compute per-word surprisal as the sum of subword surprisals, when applicable.
Additionally, punctuation is included in these estimates, although see \cref{app:results} for results omitting punctuation, which are qualitatively the same. More details are given in \cref{app:data}.

\newcommand{\testdata}{\mathcal{D}}
\paragraph{Evaluation.}
Following \citet{wilcox2020} and \citet{meister+al.emnlp21a}, we quantify the predictive power of a variable of interest as the mean difference in log-likelihood $\deltaLL$ of data points under models with and without access to that variable. 
A positive $\deltaLL$ value indicates the model with this predictor fits the observed data more closely than a model without this predictor. 
We use 10-fold cross-validation to compute $\deltaLL$ values, taking the mean across the held-out folds as our final metric. 
Our baseline model for predicting RTs contains predictors for surprisal, unigram log-frequency, character length, and the interaction of the latter two. 
These values, albeit computed on the previous word, are also included to account for spill-over effects \cite{smith2013-log-reading-time}. Surprisal from two words back is included for SPR datasets. Unless otherwise stated, GPT-2 estimates are used for baseline surprisal estimates in all models.\looseness=-1

\paragraph{Results.}

Here we explore the additional predictive power that $\infcontentk$ gives us when modeling clause-final RTs. 
In \cref{fig:LL_full}, we observe that often the additional information provided by $\infcontentk(\ww)$ indeed leads to better models of clause-final RTs.  Note that the estimated coefficients for $\infcontentk$ are always positive when $\deltaLL > 0$ (see \cref{app:coefs}), suggesting that higher values of $\infcontentk(\ww)$ correspond to longer wrap-up times. This finding is in line with other information-theoretic analyses of RTs (discussed in \cref{sec:inc}), which have consistently found positive relationships between information content and RT.

In most cases, $\infcontentk$ at some value of $k>0$ leads to larger gains in predictive power than $k=0$. 
Ergo, the information content of the preceding text is more indicative of wrap-up behavior than length alone. 
Further, while often within standard error, $\infcontentk(\ww)$ at $k>1$ provides more predictive power than at $k=1$ across the majority of datasets. 
This indicates that unevenness in the distribution of surprisal is stronger than the total surprisal content alone as a predictor of clause-final RTs.
The same experiments for sentence-medial words show these quantities are less helpful when modeling their RTs. Note that these effects hold above and beyond the spill-over effects from the window immediately preceding the sentence boundary. The effect of the distribution of surprisal throughout the sentence is stronger for eye-tracking data than for SPR; further, the trends are even more pronounced when measuring \emph{regression times} for eye-tracking data (see  \cref{app:results}).

Notably, we see some variation in trends across datasets. Due to the nature of psycholinguistic studies, it is natural to expect some variation due to, e.g., data collection procedures or inaccuracies from measurement devices. Another (perhaps more influential) factor in the difference in trends comes from the variation in dataset sizes. We see that with the smaller datasets (e.g., UCL and Provo), there may not be enough data to learn accurate model parameters. This artifact may manifest as the noisiness or a lack of a significant increase in log-likelihood (on a held-out test set) over the baseline that we observe in some cases. 

When considering prior theories of wrap-up processes, 
these results have several implications. For example, they can be interpreted as supporting and extending \citeposs{rayner2000} hypothesis, which suggests the extra time at sentence boundaries is spent resolving prior ambiguities. In this case, the observed correlation between wrap-up times and $\infcontentk(\ww)$ may potentially be linked to two factors: (1) contextual ambiguities increasing variation in per-word information content, and (2) contextual ambiguities being resolved at clause ends. On the other hand, these results provide evidence against the hypothesis that the cognitive processes occurring during the comprehension of sentence-medial and clause-final words are the same. Further, it also goes against \citeposs{HIROTANI2006425} hypothesis (discussed in \cref{sec:wrap}), as the differences in sentence-medial and clause-final times cannot be purely quantified by a constant factor.\looseness=-1

\section{Conclusion}

We attempt to shed light on the nature of wrap-up effects by exploring the relationship between clause-final RTs and information-theoretic attributes of text. We find that operationalizations of the information contained in the preceding context lead to better predictions of these RTs, while not adding significant predictive power for sentence-medial RTs. 
This suggests that information-theoretic attributes of text can shed light on the cognitive processes happening during the comprehension of clause-final words. Further, these processes may indeed be different in nature than those required for sentence-medial words. 
In short, our results provide evidence (either in support or against) about several theories of the nature of wrap-up processes.\looseness=-1

\section*{Ethics Statement}
All studies involving human evaluations were conducted outside of the scope of this paper. The authors foresee no ethical concerns with the work presented in this paper.

\section*{Acknowledgments}
RC acknowledges support from the Swiss National Science Foundation (SNSF) as part of the ``The Forgotten Role of Inductive Bias in Interpretability'' project.
TP is supported by a Facebook Fellowship Award. RPL acknowledges support from NSF grant 2121074.

\bibliography{anthology,custom}
\bibliographystyle{acl_natbib}

\newpage
\clearpage
\appendix

\section{Experimental Setup}\label{app:data}
\subsection{Data Pre-processing}
We use the Moses decoder\footnote{\url{http://www.statmt.org/moses/}} tokenizer and punctuation normalizer to pre-process all text data. Some of the Hugging Face tokenizers for respective neural models performed additional tokenization; we refer the reader to the library documentation for more details. We determine clause-final words as all those ending in punctuation. Capitalization was kept intact albeit the lowercase versions of words were used in unigram probability estimates. We estimate unigram log-probabilities on WikiText-103 using the KenLM \cite{kenlm} library with default hyperparameters. We removed outlier word-level reading times (specifically those with a $z$-score $>3$ when the distribution was modeled as log-linear). 

\subsection{Surprisal Estimates}

We use pre-trained neural language models to compute most surprisal estimates. For reproducibility, we employ the model checkpoints provided by Hugging Face \cite{wolf-etal-2020-transformers}. Specifically, for GPT-2, we use the default OpenAI version (\texttt{gpt2}); for TransformerXL, we use a version of the model (architecture described in \citet{dai-etal-2019-transformer}) that has been fine-tuned on WikiText-103 (\texttt{transfo-xl-wt103}); for BERT, we use the \texttt{bert-base-cased} version. Notably, BERT models the probability of a word given both prior and \emph{later} context, which means it can only give us pseudo estimates of surprisal. Both GPT-2 and BERT use sub-word tokenization.   We additionally use surprisal estimates from a 5-gram model trained on WikiText-103 using the KenLM \cite{kenlm} library with default hyperparameters for Kneser--Essen--Ney smoothing.

\newpage

\section{Additional Results}\label{app:results}

\begin{figure}[h]
         \centering
         \includegraphics[height=0.5\linewidth]{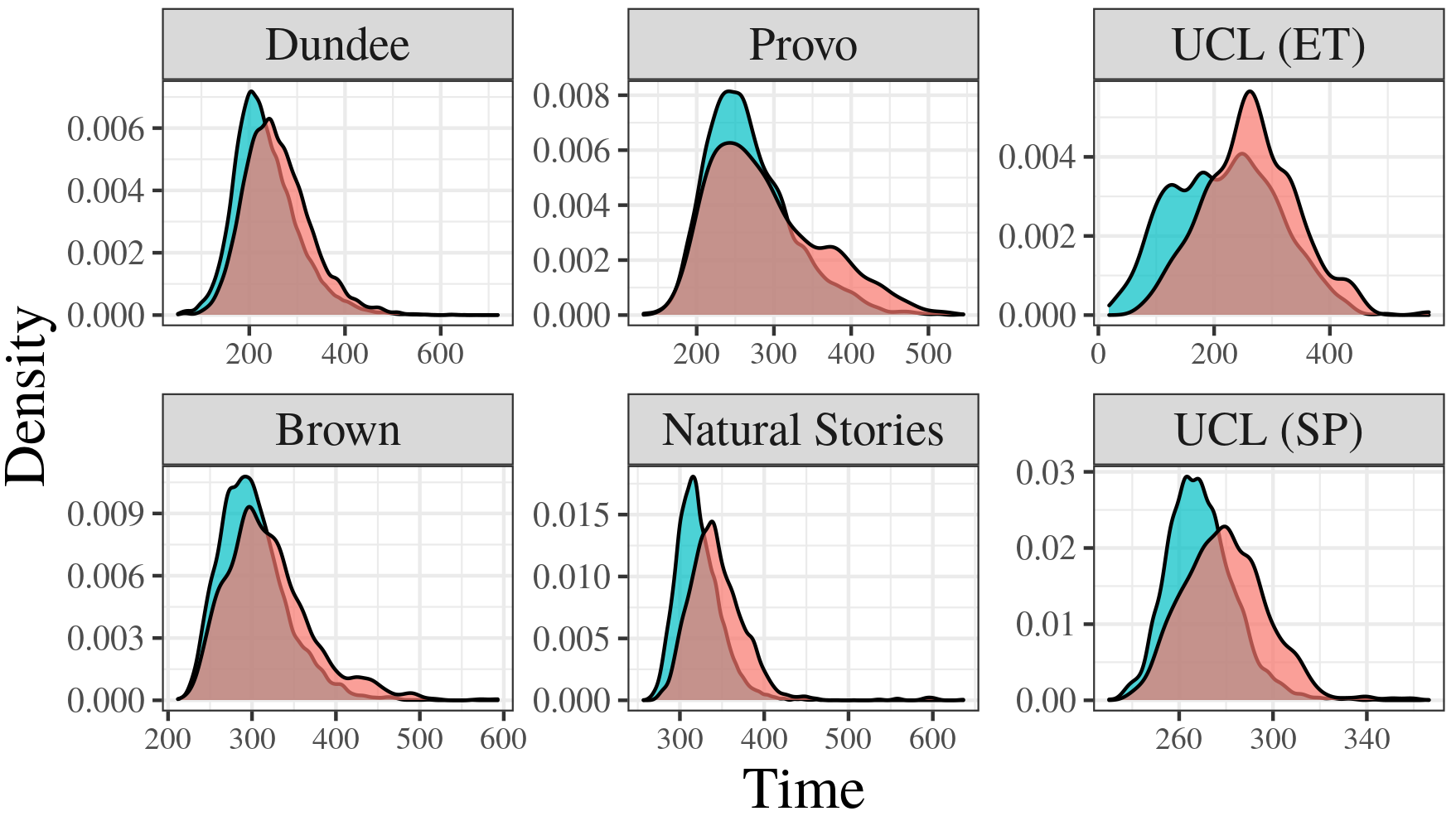}

        \caption{Distributions of average RTs for clause-final and non-clause-final words. Outlier times (according to log-normal distribution) are excluded from averages for both graphs.  
        The top-level datasets contain eye-tracking data while the bottom contain SPR data.\looseness=-1}
        \label{fig:densities_clause}
\end{figure}

\begin{figure}[H]
    \centering
    \includegraphics[width=\linewidth]{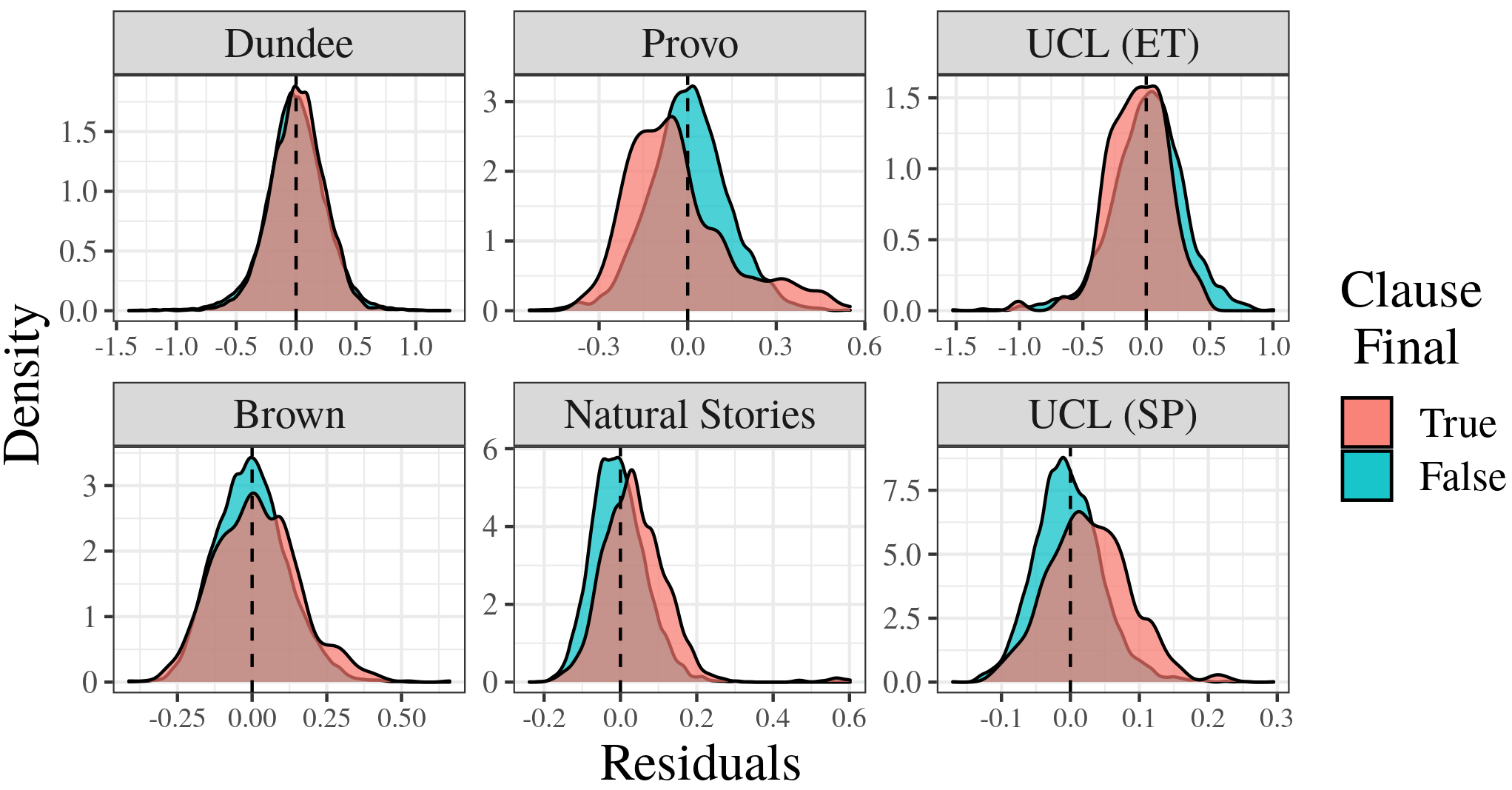}
    \caption{Version of \cref{fig:densities} where surprisal estimates do \emph{not} include the surprisal assigned to punctuation, which is often a large contributor to clause-final surprisal estimates. We see very little qualitative difference with \cref{fig:densities}.
    \vspace{5pt}}
    \label{fig:no_punct}
\end{figure}

\subsection{Regression Times Analysis}
\begin{figure}[!htb]

     \centering
     \begin{subfigure}{0.45\textwidth}
         \centering
         \includegraphics[width=\linewidth]{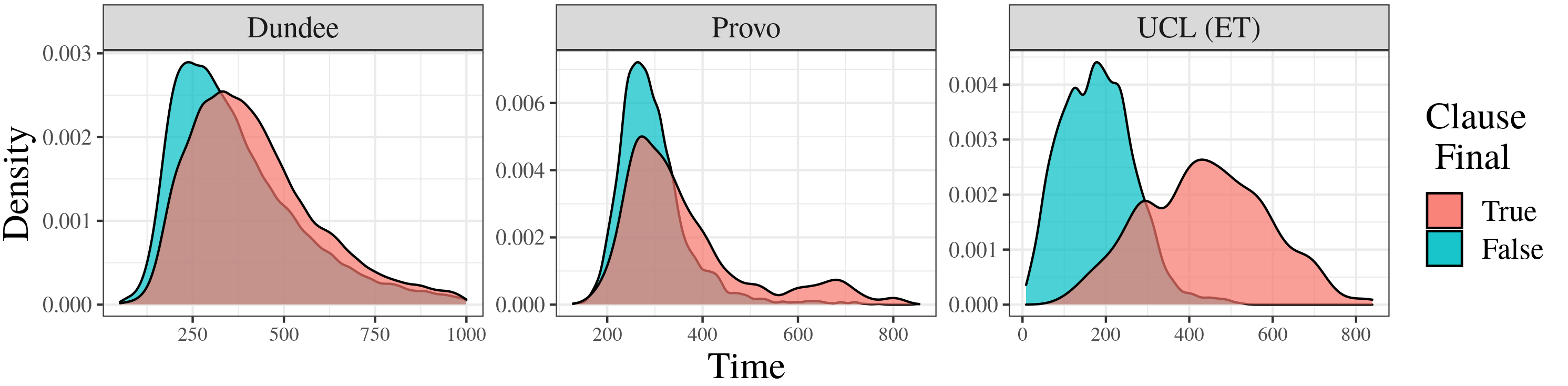}
     \end{subfigure}
     \vfill
      {\small (a)}
      \vfill
     \begin{subfigure}{0.45\textwidth}
         \centering
         \includegraphics[width=\linewidth]{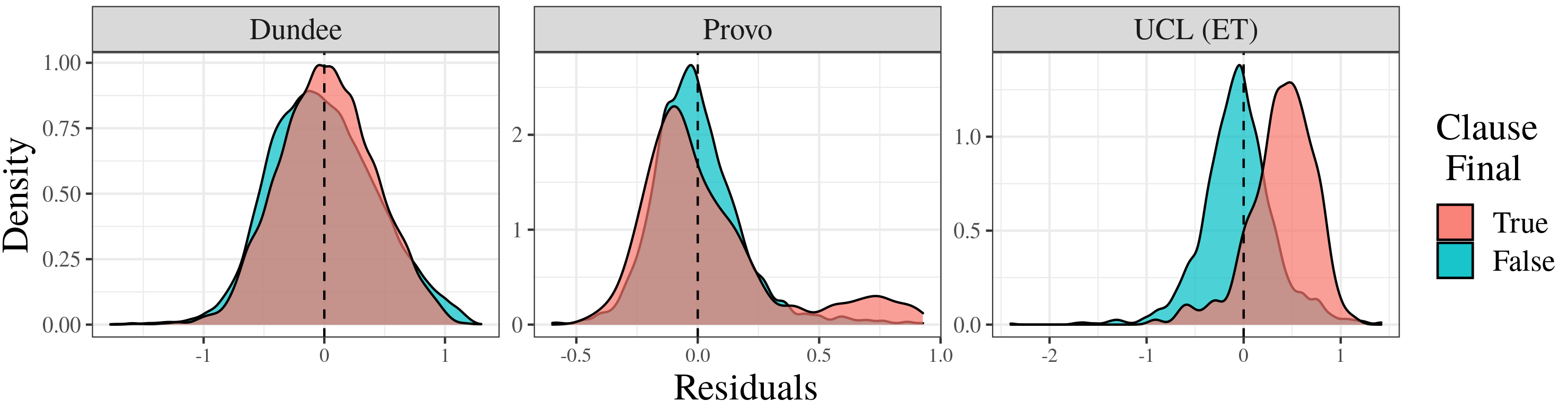}
     \end{subfigure}
     \vfill
    {\small (b)}
        \caption{Version of (a) \cref{fig:densities_clause} and (b) \cref{fig:densities} for regression times for clause-final and non-clause-final words. 
        Only applicable for eye-tracking datasets }
        \label{fig:reg_densities}
\end{figure}
\onecolumn
\begin{figure}[!htb]

     \centering
     \begin{subfigure}{0.45\textwidth}
         \centering
         \includegraphics[width=\linewidth]{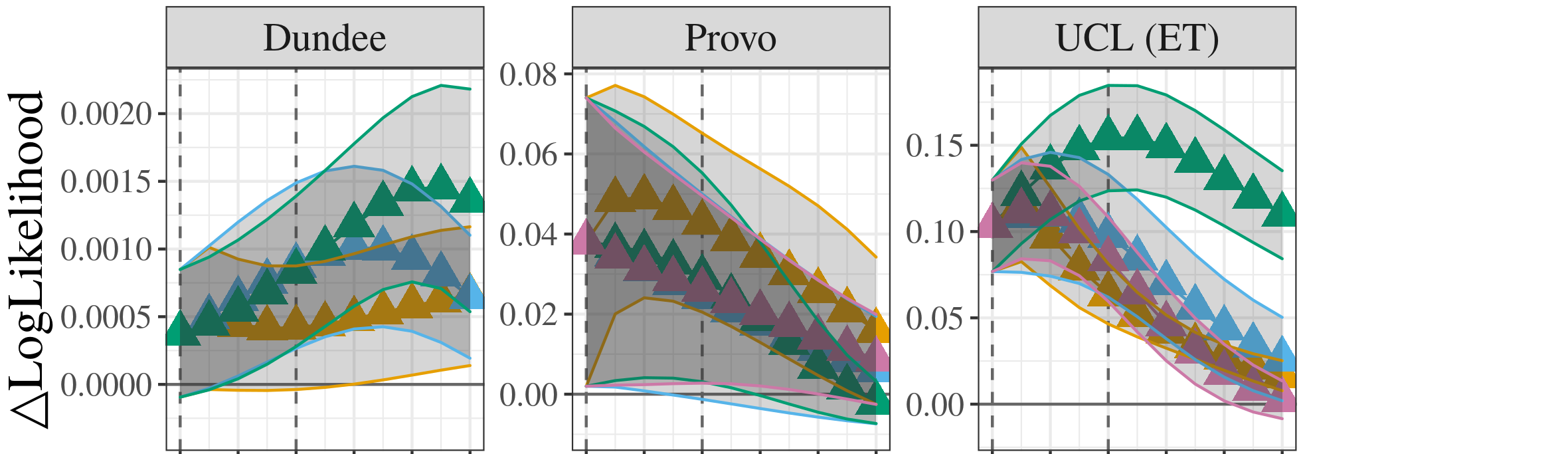}
     \end{subfigure}
     \vfill
      {\small (a)}
      \vfill
     \begin{subfigure}{0.45\textwidth}
         \centering
         \includegraphics[width=\linewidth]{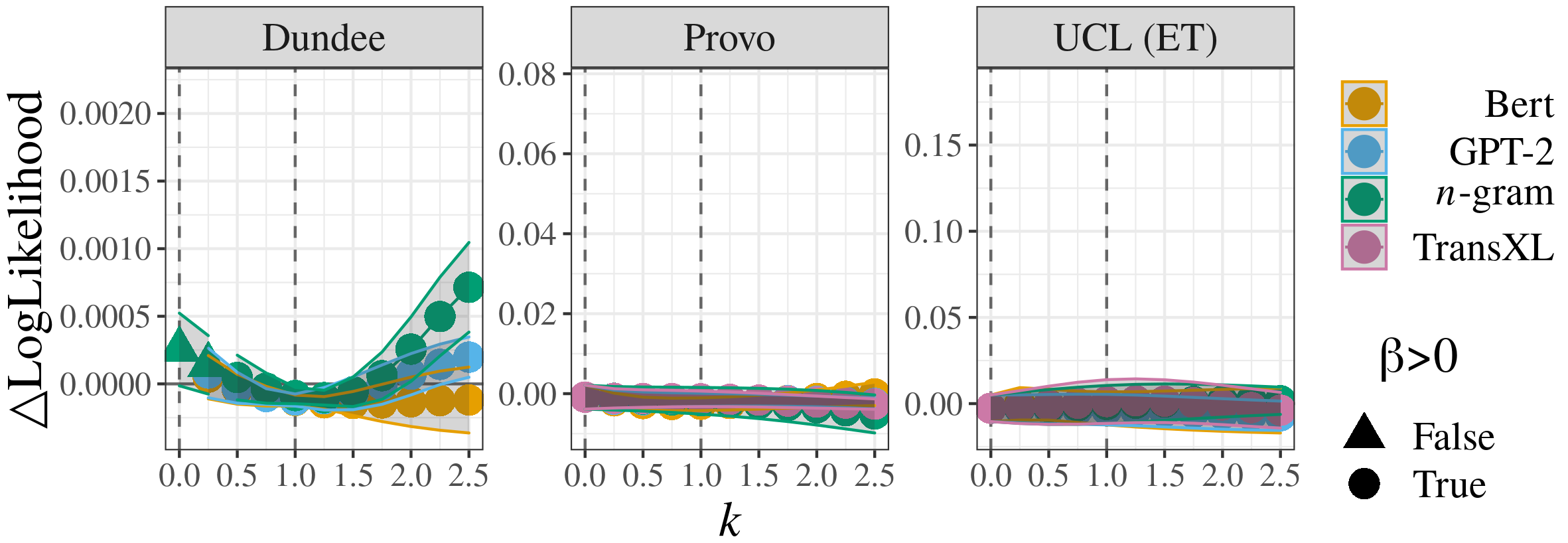}
     \end{subfigure}
     \vfill
    {\small (b)}
        \caption{Same setup as \cref{fig:LL_full} albeit predicting regression times. Results are only applicable for eye-tracking datasets. (a) shows results for predicting clause-final words, while (b) shows results for predicting sentence-medial words.  }
        \label{fig:reg_lls}
\end{figure}

\begin{figure*}[!h]
{\small (a)}
     \centering
     \begin{subfigure}{0.9\textwidth}
         \centering
         \includegraphics[width=\textwidth]{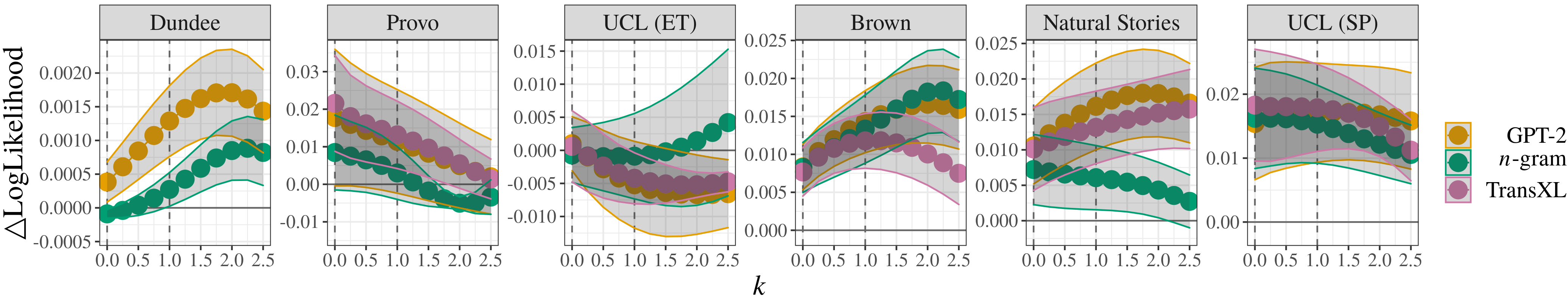}
     \end{subfigure}
      \\ {\small (b)}
     \begin{subfigure}{0.9\textwidth}
         \centering
         \includegraphics[width=\textwidth]{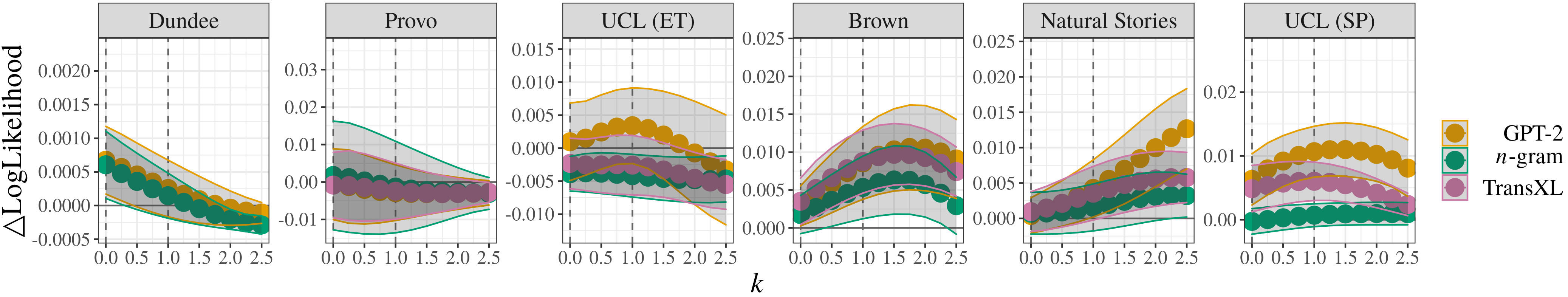}
     \end{subfigure}
    \caption{Same setup as \cref{fig:LL_full} albeit using respective model estimates for the baseline per-word surprisal estimate. (a) shows results for predicting clause-final words, while (b) shows results for predicting sentence-medial words. Results follow similar trends to those seen in  \cref{fig:LL_full}. }
    \label{fig:LL_no}
        \label{fig:LL_all}
\end{figure*}
\subsection{Predictor Coefficients}\label{app:coefs}
\begin{figure}[!htb]
         \centering
         \includegraphics[width=\linewidth]{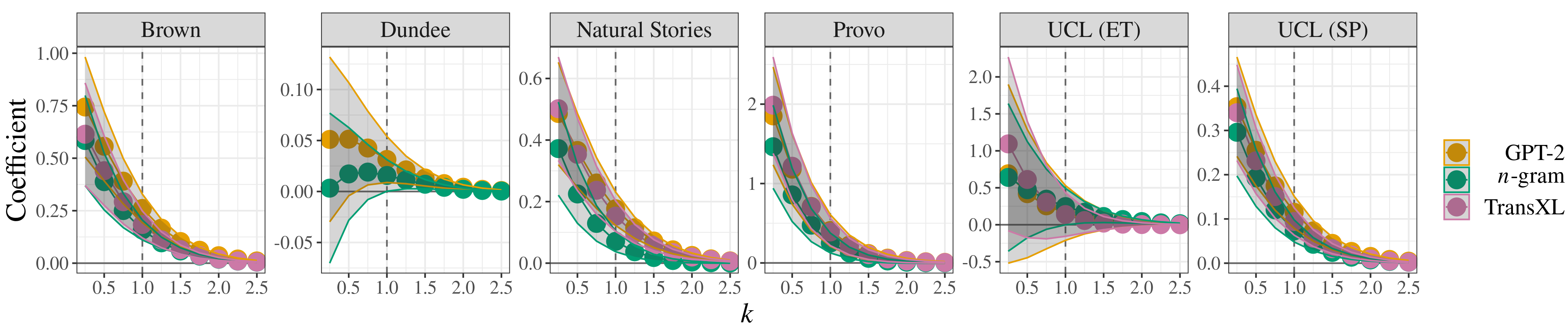}
   
        \caption{Estimated coefficients for $\infcontentk$ predictors used in in \cref{fig:LL_full} (a).  }
        \label{fig:coefs}
\end{figure}

\end{document}